# Attack on Grid Event Cause Analysis: An Adversarial Machine Learning Approach

Iman Niazazari *Student Member, IEEE*, Hanif Livani *Member, IEEE*

*Abstract*— With the ever-increasing reliance on data for data-driven applications in power grids, such as event cause analysis, the authenticity of data streams has become crucially important. The data can be prone to adversarial stealthy attacks aiming to manipulate the data such that residual-based bad data detectors cannot detect them, and the perception of system operators or event classifiers changes about the actual event. This paper investigates the impact of adversarial attacks on convolutional neural network-based event cause analysis frameworks. We have successfully verified the ability of adversaries to maliciously misclassify events through stealthy data manipulations. The vulnerability assessment is studied with respect to the number of compromised measurements. Furthermore, a defense mechanism to robustify the performance of the event cause analysis is proposed. The effectiveness of adversarial attacks on changing the output of the framework is studied using the data generated by real-time digital simulator (RTDS) under different scenarios such as type of attacks and level of access to data.

*Index Terms*—Adversarial machine learning, convolutional neural network (CNN), defense mechanism, event cause analysis, real-time digital simulator (RTDS)

## I. INTRODUCTION

### A. Background

With the proliferation of advanced measurement devices, such as phasor measurement units (PMUs) and time-synchronized transient recorders (TRs), data-driven monitoring, operation, and control tools are becoming more viable for grid operators. Event diagnostics is one of the important monitoring tools that is crucial to increase the situational awareness of abnormal conditions, such as faults, to schedule remedial actions and preventive maintenance of the critical asset, and to avoid unexpected outages. Nowadays with the increase in the use of machine learning (ML) algorithms in several power systems applications such as event cause analysis, the security and robustness of such methods have become an important research issue. Event cause analysis and root cause analysis play substantial roles for event diagnostics in power grids [1]-[5]. However, ML-based event cause analysis frameworks are prone to adversarial data manipulation attacks [6]-[13]. Adversaries can attack the data to mislead the classifier about the cause or location of events, and consequently, make the grid suffer economically or technically. Hackers can manipulate the data transmitted to the control center; they can also compromise computers that carry out the analysis. This paper is motivated by the need on how to make ML-based event cause analysis, robust against adversarial and undetected attacks. This paper presents an adversarial machine learning approach for vulnerability assessment of transient events cause analysis in power grids. In this paper, we consider adversarial machine learning (AML) for two main reasons. First, there might be some perturbations in the measurement stream because of malicious attacks. As the perturbations might be very small, the traditional residual-based bad data detection may most likely not be able to capture the bad data, and the manipulated data is used as input to the event cause analysis framework. Second, as the number of events in the real-world might not be sufficient, the dataset for training the cause analysis model is generally complemented with model-based simulation results that in many cases might have discrepancies with real-world datasets due to uncertainties such as network configuration and parameters changes. The results of this paper validate the idea of AML for creating an event analytics framework that is robust with respect to adversarial conditions or cyber-physical uncertainties.

### B. Contributions

While the concept of AML in power grids is fairly new, it has drawn some researchers' attention. In [12] the authors present a data-driven renewable generation model using a generative adversarial neural network. In [13], the authors study the vulnerability of load forecasting to data injection attacks. However, this paper offers the following unique contributions:

- Establishing a vulnerability assessment of ML-based event cause analysis with respect to the number and location of measurements under attacks, to find the most susceptible ones for an adversary to attack or for an operator to defend.
- Presenting a defense mechanism to make the event cause

Iman Niazazari and Hanif Livani are with the Department of Electrical and Biomedical Engineering, University of Nevada, Reno, NV 89557, USA
(e-mail: niazazari@nevada.unr.edu, hlivani@ieee.org)

This material is based upon work partially supported by the Department of Energy National Energy Technology Laboratory under Award Number DE-OE0000911. This work was prepared as an account of work sponsored by an agency of the United States Government. Neither the United States Government nor any agency thereof, nor any of their employees, makes any warranty, express or implied, or assumes any legal liability or responsibility for the accuracy, completeness, or usefulness of any information, apparatus, product, or process disclosed, or represents that its use would not infringe privately owned rights. Reference herein to any specific commercial product, process, or service by trade name, trademark, manufacturer, or otherwise does not necessarily constitute or imply its endorsement, recommendation, or favoring by the United States Government or any agency thereof. The views and opinions of authors expressed herein do not necessarily state or reflect those of the United States Government or any agency thereof.

analysis framework robust against stealthy data tampering attacks. Additionally, the defense mechanism is important because of potential discrepancies between real-world and model-based datasets because of uncertainties in the network parameters or measurement noise.
- Validating the proposed approach using the datasets generated using real-time digital simulator (RTDS).

The rest of this paper is organized as follows. Section II describes the adversarial machine learning event cause analysis framework. In Section III, the results are presented. Finally, Section IV presents the conclusion and future work.

## II. ADVERSARIAL ATTACK ON EVENT CAUSE ANALYSIS

There are several ways of cyberattacks, such as denial of service (DoS), or false data injection attacks (FDIA). This paper focuses on one of the most challenging cyber-attacks as malicious data tampering attacks with small perturbations on voltage or current data that can remain undetected by bad data detectors. In this section, the attack models are formulated using adversarial machine learning framework.

### A. Adversarial Goal for Event Misclassification

An adversarial attack is composed of maliciously changing the voltage or current data in such a way that the variations are almost not noticeable to the human operators, and are missed by residual-based bad data detectors as well. The altered voltage or current dataset is called adversarial data, and when fed to event classifier would be misclassified, while the original one is correctly classified. The attacker, in our study, attempts to force the target model to misclassify inputs in any class different from their true class. In other words, the adversary tries to craft a new modified version of the input $\vec{x}$, named adversarial data, and denoted as $\vec{x^*}$, misclassified by the original ML model T: $T(\vec{x^*}) \neq T(\vec{x})$. In this paper, the adversarial data (i.e., voltage) will be obtained by solving the following optimization problem:

$$\vec{x^*} = \vec{x} + \arg\min\{\vec{z}: T(\vec{x} + \vec{z}) \neq T(\vec{x})\} = \vec{x} + \delta_{\vec{x}} \quad (1)$$

where $X$ is the matrix of voltage streams from measurement locations over the sampling period, $\delta_{\vec{x}}$ is a small perturbation that is added to the original matrix of voltage streams and can still be unnoticeable as the adversary designs it to avoid being captured by the bad data detector. Doing so, it forces the classifier to be fooled and the system operators to make wrong decisions, and therefore, the power quality and root cause analysis, and preventive maintenance scheduling frameworks are not trusted.

### B. Type of Adversarial Attack

In general, there are two types of attacks that adversaries can implement; white-box and black-box attacks. In white-box attacks, the adversaries have access to the ML model's parameters, such as architecture, number, type, and size of layers. In black-box attacks, the adversaries have no access to the aforementioned parameters. In other words, the adversaries craft the adversarial manipulation using its own model which might be a different model compared to the original model or maybe crafted with no model at all. In this type of attack, which is much harder compared to the white-box attack, adversaries do not need to know the details of the ML-based event cause analysis framework to compromise it. In cyber-physical power systems, there are different layers of security which makes it difficult for adversaries to have access to the parameters of ML-based frameworks inside utilities' control rooms. This is another reason to consider black-box attacks which rely on only having access to a small portion of the voltage and current dataset.

### C. Adversarial Data Repository

It is assumed that the adversary has been monitoring the power grid for enough period of time or compromised computers inside utilities, and has gathered information about past events. Using an unsupervised clustering algorithm, the adversary can create different clusters of potential events. Based on the obtained clusters, the adversary categorizes the events into different labeled classes. Using the labeled dataset, the adversary makes its own substitute ML-based classifier for launching future attacks on the system.

### D. Substitute Model for Adversaries

As mentioned before, based on a small repository of the collected dataset, the adversary seeks to make its own model, called substitute model, and to launch attacks by making small undetectable perturbations on voltage measurement streams. The goal of the substitute model is not to achieve the optimal accuracy but to learn similar decision boundaries as the original event cause analysis framework and make a close approximation of it. The adversarially manipulated voltage stream would then have the best chance of success for fooling the classifier used inside the target utility.

The question that comes to mind is "*How can the adversary select the architecture of the substitute model when it doesn't know anything about the architecture of the target model?*"

In fact, the structure of the original classifier is not an obstacle for the adversary since it has some knowledge about it in terms of the input-output relationship. For instance, it is known that the original classifier's inputs are voltage or currents streams, and the outputs are the type of events. Therefore, the adversary can create an approximate structure for the substitute classifier. For the event cause analysis problem, the convolutional neural network (CNN) is suitable since it incorporates the spatiotemporal feature representations of events [14]. In addition, it is shown in [6] that the architecture and parameters of deep neural network substitute models have an insignificant impact on succeeding the adversarial goal. Furthermore, it is always possible for the adversary to examine a large set of architecture to find the best performance for crafting attacks. Once the suitable substitute classifier is selected, the adversary uses the collected dataset to train the model for making adversarial data manipulation

### E. Adversarial Data Crafting

After training the substitute classifier model, the adversary uses it to generate adversarially manipulated voltage samples. In this paper, we implement an approach introduced in [10] by Goodfellow et al, namely, fast gradient sign method (FGSM) for crafting the adversarial samples. In this method, given a model $S$ with the cost function of $c(S, \vec{x}, y)$, where $\vec{x}$ is the original data (i.e., voltage matrix over time and location), $y$ is the original label corresponding to the original data, the adversary can generate an adversarial sample, $\vec{x^*} = \vec{x} + \delta_{\vec{x}}$ by adding the following perturbation:

$$\delta_{\vec{x}} = \varepsilon \, sign(\nabla_{\vec{x}} c\,(S,\vec{x},y)) \quad (2)$$

where $\varepsilon$ is the value of the input parameter that controls the perturbation's value. This value is selected such that the residual-based bad data detector fails to flag it as a potential bad data. As the value of $\varepsilon$ increases, the chance of the adversarial input being misclassified by the original classifier increases, while it also increases the chance of being detected by operators or the bad data detector. It should be noted that while the adversary perturbs the input, it tries to stay stealthy as well. Therefore, the following constrained optimization problem is solved:

$$\min \|\delta_{\vec{x}}\|_p$$
$$\text{w.r.t. } |\vec{x} + \delta_{\vec{x}}| < 1 \quad (3)$$

where $\|.\|_p$ is the p-norm.

In other words, the adversary attempts to compute the loss function's gradient to decide in which direction the values must be perturbed. After making the substitute model trained and tested, the adversary intercepts voltage streams before being used by the event classifier at the control center to perturb them in a way that the classifier misclassifies the real underlying event as a different event.

### III. RESULTS AND DISCUSSIONS

In this section, the case study, results, and discussions about events cause analysis are presented.

#### A. Test Case

To assess the validity of the proposed methodology, the WSCC 9-bus system [15] is implemented on the real-time digital simulator (RTDS) facility at the University of Nevada, Reno (UNR). Three software-based TRs are located at bus 4, 7, and 9 which measure three-phase voltages at these buses. The simulation time step is 50 µsec corresponding to 20 kHz sampling rate. The measurements are recorded for two cycles ($2/60 \approx 34$ ms, 34 ms× 20 kHz = 680 samples) before any protection devices operate. This duration is determined based on extensive studies of different events. This is also practically desirable as high-frequency measurements streaming can start once an event is detected by the device, and stop after a pre-determined period, due to the very large communication bandwidth required for transmitting such large datasets. Such a concept exists in line monitoring devices manufactured by Sentient Energy [16].

#### B. Events Descriptions

In this paper, the following three events in power transmission grids are modeled. The simulated events cover a diverse set of signatures, such as type, the location, inception angles, and different values that can occur in power grids.

- *Class 1 (Line Energization):* This class is simulated by closing one end of the line at the time $t_0$ while the other end of the line is already closed. The switching time ($t_0$) varies from 8 ms to 23 ms with a step of 1 ms (covering the whole period of voltage waves, $1/60 \approx 17$ ms). This experiment is repeated for each switch resulting in 6×2×16=192 experiments. For synchronized operation of circuit breakers at both ends of a line, first the sending end of the line at $t_0$ is closed and the receiving end is closed at $t_0 + \Delta t$, where $t_0$ changes from 8 ms to 23 ms and $\Delta t$ changes from 0 msec to 3 ms with one ms intervals. The delay ($\Delta t$) is added to take the practical consideration of synchronization into account. This experiment is repeated for each line resulting in 6×4×16=384 experiments. Therefore, the total experiment in this class is 576.

- *Class 2 (Capacitor Bank Energization):* Energization inrush is a transient occurring when the capacitor at the bus is energized. This class is simulated based on switching of the capacitor banks on 3 buses. For the capacitor bank energization, it is assumed three capacitors are located at buses 5, 6, and 8 with capacities of 25, 50, 75, and 100 MVAR. In addition, the simulations carried out in 16 different inception angles to account for different intensity of switching. The total number of experiments in class 2 is 3×4×16=192.

- *Class 3 (Fault):* This class covers all the faults in the system, including single-phase, two-phase, and three-phase. This class embodies a wide range of parameters including fault resistance, locations, and inception angles. It is assumed the faults happen every 20% of the lines, therefore, four locations on each line. Three inception angles are considered to account for different switching moments. The fault events are simulated for fault resistor values of 0.1 and 5 ohms. Therefore, the total number of simulated experiments for this class is 6×3×4×4×2=576.

#### C. Machine Learning-Based Event Cause analysis

The original three-phase voltages are transformed into the modal domain and mode-1 voltage (i.e., the aerial mode) is fed as the input to the CNN-based event cause analysis to identify the cause of an event. It should be noted that for each simulated event, all the mode-1 voltages from the sensor locations are stacked on top of each other to form a spatiotemporal measurement matrix which is then converted into a 2D image for CNN training and classification. An extensive number of training and evaluation scenarios are performed to discover the most suitable CNN parameters, i.e., the number of convolutional layers, fully connected layers, number and size of the filters, etc. The optimal CNN structure is as follows: one convolutional layer with ten 2×20 filters, the stride of 2, one ReLU for the activation layer, and one fully connected layer, with softmax layer for the classification as the final stage. The classifier is trained on a single CPU using the Adam optimizer with the number epoch of 6, batch size of 8, and the initial learning rate of 0.001.

#### D. Adversarial Data Generation

Fig. 1 (a) illustrate the adversarial samples generated using the FGSM algorithm. It shows how the voltage signals originally corresponding to an event can be altered to force a classifier to misclassify. In each image of Fig. 1, the y-axis represents the location of voltage measurement (three buses), and the x-axis represents the number of samples (680 samples). It needs to be noted that the voltage streams are normalized from 0 to 1. The images on the diagonal locations show the original voltage streams, while the off-diagonal images show the adversarially perturbed voltage samples. For instance, the second image in the first row of Fig. 1 (a) shows that by adding a small perturbation to the original voltage samples, the classifier is forced to misclassify the event as class 2, while the image is not undergone a noticeable alteration, and can still remain undetected by the conventional bad data detector. In addition, to compare the level of stealthy of adversarial voltages generated by the FGSM algorithm, another type of attack, namely, Jacobian-based Saliency Map Attack (JSMA) [8], is used to generate perturbed voltage streams as shown in Fig. 1. (b). As it can be seen, the adversarially generated voltage samples, in the second case, is very noticeable that can be detected by operators or by the bad data detector.

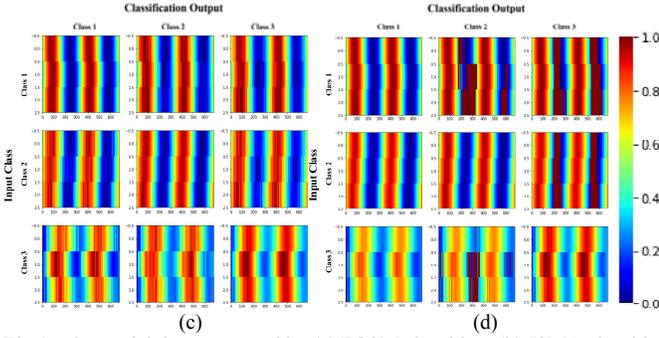

Fig.1. adversarial data generated by (a) FGSM algorithm, (b) JSMA algorithm

### E. Evaluation of Adversarial Attacks Success

The total number of 1344 events is divided into 940 (70% of the dataset) training set, and 204 ($\approx$15% of the dataset) testing set, to evaluate the original classifier. The remaining 200 events are divided into 150, and 50 for training and testing of the substitute model used by the adversary. The architecture of the substitute model is as follows. Five filters of 2×10, with the stride of 1, one ReLU for the activation layer, one fully connected layer, and softmax layer for the classification. The accuracies for the original classification model and substitute model are calculated as 99.50% and 94%, respectively. To evaluate the effects of adversarial voltage samples on the event cause analysis performance, two metrics are introduced: success rate, and transferability. Success rate is the percentage of the adversarial voltages that are misclassified by the model. The transferability is the percentage of the adversarial voltages crafted according to the substitute model and are misclassified by the original classifier. The transferability is more important since the final goal of any adversary is to force the original classifier to misclassify the tampered data without being detected. Fig. 2 (a) shows the success and transferability rates for different values of perturbation coefficient ($\varepsilon$) with the infinite-norm in Eq. (3). As it can be seen, as $\varepsilon$ increases, the success and transferability rates also increase. That is because an increase in the value of $\varepsilon$ increases the perturbation on the original voltage stream and therefore results in forcing the original classifier to misclassify the adversarially crafted data. It can be seen that for $\varepsilon$=0.07 the adversarial data are able to fool the target model by 76% which is very effective. In addition, based on the closeness of success rate and transferability, it is obvious that the substitute classifier performs well in approximating the decision boundaries of the original classifier. However, bigger values of the perturbation coefficient increase the chance of being detected by the bad data detector.

### F. Impact of Attack Order on Attack Success

As we can recall from Eq. (3), to create the adversarial voltage streams, the perturbations can be optimized with respect to different norms. The FGSM attack is originally optimized for infinite-norm as it is implemented in this paper [7]. However, to evaluate the effect of different norms on adversarial voltage generation, two other norms are studied: 1-norm and 2-norm. Fig. 2 (b) shows a comparison between these three norms with different perturbation levels. As it can be seen, in attacks with infinite-norm, the transferability rate of adversarial voltages increases with the increasing perturbation values. However, the attacks based on 1-norm and 2-norm are not successful and the transferability rate stays below 5%.

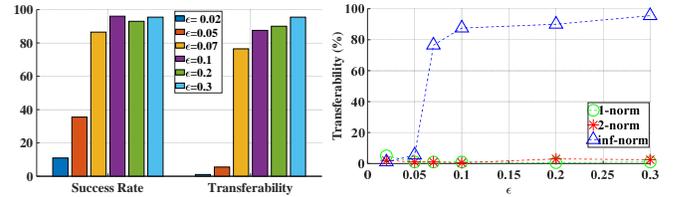

Fig.2. (a) Success rate and transferability of adversarial data with respect to $\varepsilon$, (b) Transferability of adversarial data for different norm orders

### G. Impact of Level of Access to Data by Attackers

Another important issue is the amount of dataset available to adversaries. The amount of data available to the adversary can be a function of time, the extent of cybersecurity breaches, stealthy level, and usefulness of previously collected data for clustering the events. In this paper, the dataset available to train the adversarial substitute classifier is assumed to be 25, 50, 100, 125, and 150 events. As can be seen in Fig. 3 (a), as the number of datasets increases, the adversary is more likely to force the original classifier to misclassify the adversarially tampered voltage stream. This is because of the closeness of the substitute model to the original classifier.

### H. Vulnerability with Respect to the Number of Compromised Measurements

In this paper, the vulnerability of the CNN-based event cause analysis with respect to the number of compromised measurements is studied. Generally, an adversary may not have access to all the streaming measurements or past events. This can be mostly due to strong cybersecurity measures inside substations or utilities. Therefore, the vulnerability analysis allows the adversary to spend its time and money on more vulnerable substations and still perform a successful attack on the event cause analysis framework. On the other hand, the vulnerability analysis is of interest to system operators, since they can decide to reinforce more susceptible substations by better defending them against adversaries. To simulate the vulnerability assessment on k nodes, it is assumed that if the input voltage matrix is $n \times t$, the adversary has only access to $k$ of n measurements. Therefore, the adversarial voltages crafted by the substitute model will be a $k \times t$ voltage matrix which will be combined with the rest of $n - k \times t$ "clean" data and used by the original classifier. Fig. 3 (b) shows the result of the vulnerability assessment for three measurement locations on the test case. The attacks are made by the perturbation value of $\varepsilon = 0.1$. The first observation is that as the number of the attacked locations increases, the adversary is more successful in forcing the original classifier to misclassify the voltage streams with adversarial samples inside them. In addition, when the adversary attacks all the nodes, the attack is significantly more effective; however, more likely to be detected, Another observation is that attacks on buses close to the edge of the grid (e.g., bus 4) result in less successful misclassification.

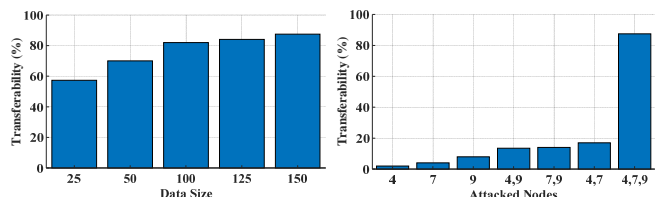

Fig. 3. (a) Transferability of adversarial data with respect to the size of available adversary dataset (b) Vulnerability assessment based on the compromised nodes

## I. Defense Mechanism

Finally, we look at the problem from the operator's point of view. Although the adversary can always make malicious attacks to force the original classifier to misclassify the stream of voltages with tampered samples, the defender can also make the original classifier more robust by finding the weak points and fortifying it to reduce the chances of misclassification. In this paper, we proceed with one of the most popular defense mechanisms, namely adversarial training, for reinforcing the original classifier and re-evaluating the effect of adversarial voltage samples. It is demonstrated that retraining the model using the adversarial samples can significantly increase the robustness of CNN-based event classifiers [9], [10].

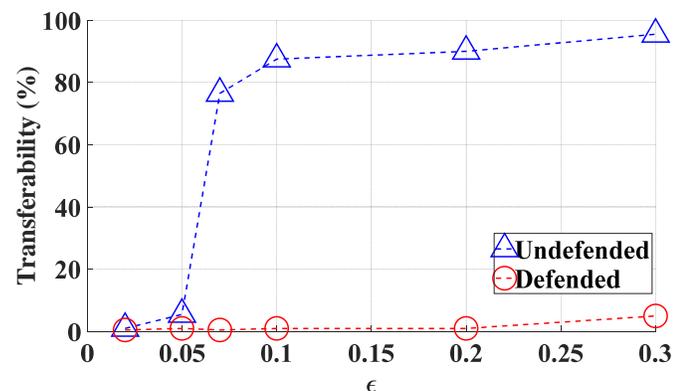

Fig.4. Transferability for the defended target model vs original target model

The adversarial voltages are generated for all 940 training sets and are augmented into the training set for retraining the original classifier. The original classifier is retrained with the 1880 dataset and tested on the adversarial data sets crafted by the adversary using the substitute model. Fig. 4 shows the misclassification rate of adversarial voltage streams created by the adversary for different perturbation parameters on both the original and retrained classifiers. As it can be seen from the figure, after retraining the original classifier by injecting the adversarial, the adversarial data made by the adversary have little impact on the classification output of the original classifier, only 5% even with $\varepsilon = 0.3$ which is a very large value, and can be easily detected by the bad data detector.

## IV. CONCLUSION AND FUTURE WORK

In this paper, an adversarial attack model on power grids' event cause analysis frameworks is presented. It is shown that data-driven event cause analysis in power grids are vulnerable to adversarially crafted data sets that even bad data detectors may fail to detect them. This paper also presents a vulnerability assessment to find the weak substations from adversarial attackers' or operators' perspective. Moreover, a defense mechanism for robustifying the original convolutional neural network-based event cause analysis framework is presented. As future work, we will implement the proposed methodology on larger networks with more events considering the effect of different sampling rates. Exploring other attacks such as Carlini and Wagner [7] and Deepfool [11] will be investigated. Finally, the robustness of the system with different defense mechanisms such as defensive distillation will be studied.

REFERENCES

[1] I. Niazazari and H. Livani, "A PMU-data-driven disruptive event classification in distribution systems," Electric Power Systems Research, vol. 157, pp. 251–260, 2018.
[2] E. Styvaktakis, M. Bollen, and I. Gu, "Expert system for classification and analysis of power system events," IEEE Transactions on Power Delivery, vol. 17, no. 2, pp. 423–428, 2002.
[3] I. Niazazari and H. Livani, "Disruptive event classification using PMU data in distribution networks," 2017 IEEE Power & Energy Society General Meeting, 2017.
[4] M. Khoshdeli, I. Niazazari, R. J. Hamidi, H. Livani, and B. Parvin, "Electromagnetic transient events (EMTE) classification in transmission grids," 2017 IEEE Power & Energy Society General Meeting, 2017.
[5] H. Khodabandehlou, I. Niazazari, H. Livani, and M. S. Fadali, "Anomaly Classification in Distribution Networks Using a Quotient Gradient System", arXiv preprint arXiv:1805.04979. 2018.
[6] N. Papernot, P. Mcdaniel, I. Goodfellow, S. Jha, Z. B. Celik, and A. Swami, "Practical Black-Box Attacks against Machine Learning," Proceedings of the 2017 ACM on Asia Conference on Computer and Communications Security - ASIA CCS 17, 2017.
[7] N. Carlini and D. Wagner, "Towards Evaluating the Robustness of Neural Networks," 2017 IEEE Symposium on Security and Privacy (SP), 2017.
[8] N. Papernot, P. Mcdaniel, S. Jha, M. Fredrikson, Z. B. Celik, and A. Swami, "The Limitations of Deep Learning in Adversarial Settings," 2016 IEEE European Symposium on Security and Privacy (EuroS&P), 2016.
[9] C. S. Szegedy, W. Zaremba, I. Sutskever, J. Bruna, D. Erhan, I. Goodfellow, and R. Fergus, "Intriguing properties of neural networks." arXiv preprint arXiv:1312.6199, 2013.
[10] I. Goodfellow, J. Shlens, and C. Szegedy, "Explaining and Harnessing Adversarial Example." ICLR, 2014.
[11] S.-M. Moosavi-Dezfooli, A. Fawzi, and P. Frossard, "DeepFool: A Simple and Accurate Method to Fool Deep Neural Networks," 2016 IEEE Conference on Computer Vision and Pattern Recognition (CVPR), 2016.
[12] Y. Chen, Y. Wang, D. Kirschen, and B. Zhang, "Model-Free Renewable Scenario Generation Using Generative Adversarial Networks," IEEE Transactions on Power Systems, vol. 33, no. 3, pp. 3265–3275, 2018.
[13] Y. Chen, Y. Tan, and D. Deka, "Is Machine Learning in Power Systems Vulnerable?," 2018 IEEE International Conference on Communications, Control, and Computing Technologies for Smart Grids (SmartGridComm), 2018.
[14] I. Niazazari, R. J. Hamidi, H. Livani, and R. Arghandeh, "Cause Identification of Electromagnetic Transient Events using Spatiotemporal Feature Learning," arXiv preprint arXiv:1903.04486, 2019.
[15] Available [online] :https://icseg.iti.illinois.edu/wscc-9-bus-system/
[16] UM3+™ Line Monitor, Available [online]: https://www.sentient-energy.com/products/um3-line-monitor